\documentclass{article}

\usepackage{PRIMEarxiv}

\usepackage[utf8]{inputenc} %
\usepackage[T1]{fontenc}    %
\usepackage{hyperref}       %
\usepackage{url}            %
\usepackage{booktabs}       %
\usepackage{amsfonts}       %
\usepackage{nicefrac}       %
\usepackage{microtype}      %
\usepackage{lipsum}
\usepackage{fancyhdr}       %
\usepackage{graphicx}       %
\graphicspath{{media/}}     %
\usepackage{multirow}
\usepackage{caption}
\usepackage{subcaption}

\title{PolyHope: Two-Level Hope Speech Detection from Tweets}

\author{
  Fazlourrahman Balouchzahi, Grigori Sidorov, Alexander Gelbukh \\
  Instituto Politécnico Nacional (IPN), Centro de Investigación en Computación (CIC), Mexico City, Mexico \\
  \texttt{\{fbalouchzahi2021, sidorov, gelbukh\}@cic.ipn.mx} \\
}

\begin{document}
\maketitle

\begin{abstract}
Hope is characterized as openness of spirit toward the future, a desire, expectation, and wish for something to happen or to be true that remarkably affects human's state of mind, emotions, behaviors, and decisions. Hope is usually associated with concepts of desired expectations and possibility/probability concerning the future. Despite its importance, hope has rarely been studied as a social media analysis task. This paper presents a hope speech dataset that classifies each tweet first into “Hope" and “Not Hope", then into three fine-grained hope categories:  “Generalized Hope", “Realistic Hope", and “Unrealistic Hope" (along with “Not Hope"). English tweets in the first half of 2022 were collected to build this dataset. Furthermore, we describe our annotation process and guidelines in detail and discuss the challenges of classifying hope and the limitations of the existing hope speech detection corpora. In addition, we reported several baselines based on different learning approaches, such as traditional machine learning, deep learning, and transformers, to benchmark our dataset. We evaluated our baselines using weighted-averaged and macro-averaged F1-scores. Observations show that a strict process for annotator selection and detailed annotation guidelines enhanced the dataset's quality. This strict annotation process resulted in promising performance for simple machine learning classifiers with only bi-grams; however, binary and multiclass hope speech detection results reveal that contextual embedding models have higher performance in this dataset.
\end{abstract}

\keywords{Hope \and Wish \and Desire \and Expectation \and Machine Learning \and Deep Learning \and Transformers \and Natural Language Processing}

\section{INTRODUCTION} \label{Introduction}

Hope is one of the exceptional human capabilities that enables one to envision future events and their possible expected outcomes flexibly. Those visions significantly affect one's emotions, behaviors, and state of mind, even though the desired outcome may have significantly less likelihood of happening~\cite{bruininks2005distinguishing}. Snyder (2000)~\cite{snyder2000handbook} considers hope a powerful compensating feature in human psychology to face challenges.

Nowadays, online social media platforms significantly affect human life, and people freely pen their thoughts on this social networks~\cite {balouchzahi2022fake}. The significant features of social media, such as rapid dissemination, low cost, accessibility, and anonymity, have increased the popularity of social media platforms~\cite{balouchzahi2021hssd}. Since social media provides deep insight into people behavior in social media, they are major sources of scientific researches on Natural Language Processing (NLP) problems~\cite{butt2021transformer}.

Hence, analyzing hope in social media is considered an essential determinant of well-being that can provide potentially valuable insights into the trajectory of goal-directed behaviors, persistence in the face of misfortunes, and the processes underlying adjustment to positive and negative life changes.

Over the last few years, several researchers explored psychological traits and other social media analysis tasks such as: emotion analysis (fear, anger, happiness, depression), hate speech, abusive language identification, and misogyny detection through NLP techniques~\cite{ashraf2021abusive}. However, hope speech on social media has rarely been explored as an NLP task. To the best of our knowledge, the Hope Speech dataset for Equality, Diversity, and Inclusion (HopeEDI)~\cite{chakravarthi2020hopeedi} is a multilingual hope speech detection corpus in English and code-mixed Dravidian languages and the corpus presented by Palakodety et al. (2020)~\cite{palakodety2020hope} in English and Hindi are the only available corpora for Hope Speech detection. 
Both the corpora model hope speech detection only as a binary Text Classification (TC) task containing two classes, “Hope" and “Not Hope". However, depending on the characteristics of hope, there may be different types of hope in texts. Hence, text belonging to the “Hope" class can be further classified as classes like “rational and irrational or realistic and wishful hopes"~\cite{garrard2009hope, wiles2008hope, verhaeghe2007does}.

Given the only two datasets for the hope speech detection task mentioned above, our dataset provides opportunities to explore further research in this domain.

In a comprehensive and general categorization, hope has been identified either as “Particularized" or as “Generalized" hope~\cite{wiles2008hope, webb2007modes, ezzy2000illness, lohne2004hope, smith2005men}. These modes of hoping are differentiated based on their objective, cognitive-affective activity, and behavioral characteristics. Particularized hope is always directed to specific outcomes and expectations, while Generalized hope lacks a concrete objective and forms open-ended expectancies towards the future~\cite{webb2007modes}. 

According to Ezzy (2000)~\cite{ezzy2000illness}; Smith and Sparkes (2005)~\cite{smith2005men}, Particularized hope is similar to the typical definition of hope used in the psychological literature as the expectation and desire for specific events and outcomes (e.g., I hope the surgery will be successful). In contrast, Generalized hope is characterized by openness to events and outcomes (e.g., I hope I will get well). In the first example, “surgery" is a specific event that one may hope to be successful, while in the second example, it seems different and only hopes for a better future. Although all types of hope represent future-oriented expectations (whether general or specific), they differ in how they influence human behaviors and decision-making ability~\cite{bailey2007hope}. Therefore, distinguishing the different constructs of hopes might be beneficial and lighten a new path in social media analysis tasks.

While there is no report on different types of Generalized hope in the literature, based on the characteristics of the desired outcomes, psychologists differentiate Particularized hope into two fine-grained sub-categories, hope as an expectation and hope as a want or wish~\cite{wiles2008hope}. Other nomenclatures for these sub-categories include : realistic and wishful hopes~\cite{eaves2016ways}; realistic and unrealistic hopes~\cite{webb2007modes, eliott2002discursive,links1994breaking}; realistic and false hopes~\cite{verhaeghe2007does}; rational and irrational hopes~\cite{garrard2009hope}.

One may hope for a high possibility outcome (e.g., we just got engaged and are hoping to get married soon) or for an outcome with the knowledge that the likelihood of its happening is remote (e.g., winning a lottery)~\cite{wiles2008hope}. Eaves et al. (2016)~\cite{eaves2016ways} distinguish Particularized hopes according to “reasonable or probable outcome, in terms of normal or expected outcomes."

Realistic hope can be characterized as a hope directed toward specific outcomes, involving a process of mental imaging along with occurrence probability calculation to prevent the person from losing his grip on reality~\cite{webb2007modes} (e.g., I have been studying a whole week, and I believe I can pass the test). In contrast, unrealistic hope is based on incomplete or incorrect information and hopes for something improbable that is not coming through~\cite{verhaeghe2007does} (e.g., I got very low marks and everyone says that I am already failed, but I am waiting for a miracle to happen). Regarding the associated behavioral response of Realistic hope, Webb (2007)~\cite{webb2007modes} argues that Realistic hope helps counteract risk aversion. Therefore, distinguishing between Realistic and Unrealistic hopes is an essential task~\cite{wiles2008hope}.

In this paper, we primarily discuss challenges, limitations, and methods used for annotation guidelines for the existing binary hope speech detection datasets, and then we present a two-level annotated hope speech detection dataset in English tweets according to the definition of hope given by the psychologists and the learning models for hope speech detection using the new dataset. Our dataset and baselines will be available on request to the corresponding author. 

The main contributions of this paper can be summarized as follows:
\begin{itemize}

    \item Study of hope speech detection as a two-level Text Classification (TC) task,
    \item Critical review of existing datasets,
    \item Developing the guidelines for annotating  binary and multiclass hope speech detection dataset,
    \item Building binary and multiclass hope speech detection dataset,
    \item Modeling hope speech detection as a multiclass classification task for the first time,
    \item Performing a range of experiments on learning approaches as baselines that provide a benchmark for future research on hope speech detection tasks.
     
\end{itemize}

\subsection{Task description}
Inspired by the earlier literature about hope, in the present study, each tweet is first classified as “Hope" or “Not Hope". Further, the “Hope" class is, in turn, fine-grained into one of three categories: “Generalized Hope", “Realistic Hope", and “Unrealistic Hope". This task consists of the following two subtasks to classify hope speech from English tweets:
\begin{itemize}
    \item \textbf{Subtask A - Binary Hope Speech Detection: }In this task, each tweet will be identified as either Hope or Not Hope, 
    \item \textbf{Subtask B - Multiclass Hope Speech Detection: }In this task, each tweet will be classified into fine-grained hope categories: Generalized Hope, Realistic Hope, and Unrealistic Hope, along with Not Hope tweets.

\end{itemize}

The rest of the paper is organized as follows:
Hope is defined in detail in~\ref{def};
Existing hope speech detection corpora, limitations, and techniques used for hope speech detection are discussed in  ~\ref{Related Wor}. RELATED WORK and steps in dataset creation are presented in~\ref{Dataset Development}. DATASET DEVELOPMENT; \ref{BENCHMARKS}. BENCHMARKS and~\ref{Results}. RESULTS describe the baselines and results, respectively, followed by the performance analysis of baselines in~\ref{Error Analysis}. ERROR ANALYSIS. Eventually,~\ref{DISCUSSION}. The DISCUSSION describes the dataset's characteristics and limitations, and we conclude the paper in~\ref{CONCLUSION}. CONCLUSION AND FUTURE WORKS.

\section{Definitions} \label{def}
Hope was studied in psychology as cognitive-based~\cite{snyder1991will} and emotion-based~\cite{averill2012rules} models. According to Snyder et al. (1991)~\cite{snyder1991will}, hope was described as a cognitive-based model and defined in terms of a goal-setting framework, where a person is motivated to remain engaged with a future outcome and can anticipate a way to reach that outcome. Conversely, Averill et al. (2012)~\cite{averill2012rules} described hope as an emotion-associated model that depends on the perceived likelihood of achieving an outcome.

There are diverse definitions of hope reported in the literature. Hope is defined as an integral part of being a human~\cite{webb2007modes, webb2013pedagogies} and usually a future-oriented thinking~\cite{lohne2004hope}. It encourages the person to transform his/her intentions to act and prevent despair and depression~\cite{garrard2009hope}.  
Verhaeghe et al. (2007)~\cite{verhaeghe2007does} describe hope as a psychological process of adapting to some unfortunate or unexpected event and situation. Eaves et al. (2016)~\cite{eaves2016ways} believe that hope is a dynamic and multi-faceted mindset that can be considered a biological and supernatural medicine that directly impacts human health. 

In the other definition, Maretha (2021)~\cite{maretha2021meaning} presents hope as personal feelings co-related with mental activities of desire and claims that it is an encouragement accompanied by desire, and the tendency that arises is in the form of real and unreal expectations.
Snyder (2002)~\cite{snyder2002hope}, more generally, describes hope as a desire for something to happen or to be true, which is commonly associated with promise, potential, support, reassurance, suggestions, or inspiration during periods of illness, anger, stress, loneliness, and depression. 

Hence, we conclude that hope is “a future-oriented expectation, desire or wish towards a general or specific event/outcome phenomenon that has a significant impact on human behavior, decision, and emotions."

\section{RELATED WORK} 
\label{Related Wor}
Hope is a partially subjective term that both psychologists and philosophers are struggling to define it~\cite{snyder2002hope}. Hope analysis can be located within social media tasks' growing interest. However, most of the ongoing research on social media is focused on controlling and eliminating harmful content such as hate speech, abuse and offensive, misogyny detection, and false information or emotion analysis tasks~\cite{chakravarthi2020hopeedi}. Hope speech detection as a Natural Language Processing (NLP) task was introduced by Chakravarthi et al. (2020)~\cite{chakravarthi2020hopeedi} and Palakodety et al. (2020)~\cite{palakodety2020hope} by proposing two multilingual corpora that classify each YouTube comment into Hope and Not Hope categories. Details of these corpora are presented in Table~\ref{tab:HopeSpeechDatasets}. The existing hope speech corpora and their limitations, followed by the techniques for hope speech detection, are described below:

\subsection{Hope speech detection corpora} \label{Hope speech detection corpora}

War-torn regions reveal a lot about the sentiments of people suffering and striving for peace. A comprehensive report~\cite{palakodety2020hope} on Kashmir (disputed territory) revealed instances of hope speech in YouTube comments after the Pulwama terror attack on February 14, 2019. Palakodety et al. (2020)~\cite{palakodety2020hope} constructed a multilingual dataset of YouTube comments in English and Hindi written in Roman and Devanagari scripts, respectively. They used a combination of polyglot embeddings from FastText (100-dimensions), sentiment score, and n-grams (1-3) with Logistic Regression (LR) to achieve the best averaged-macro F1-score of 78.51 ($\pm$ 2.24\%). They modeled hope speech detection as a positive comment mining task (positive/negative sentiments) which shows a very shallow understanding of hope as a subject. In reality, hope is a broad phenomenon with various emotions. 

Chakravarthi et al. (2020)~\cite{chakravarthi2020hopeedi} ignited the other spark of hope speech detection in social media platforms  by developing a HopeEDI corpus from YouTube comments in Dravidian and English languages. Initially, the corpus consisted of English and code-mixed Tamil-English and Malayalam-English datasets. Later, Chakravarthi et al. (2022)~\cite{chakravarthi2022overview} extended the work for Spanish and code-mixed Kannada-English texts\footnote{Malayalam, Tamil, and Kannada are Dravidian languages widely used in India}.

Similar to Palakodety et al. (2020)~\cite{palakodety2020hope}, the authors modeled the hope speech detection task as a binary classification task where each YouTube comment was identified as Hope or Not Hope. The HopeEDI corpus is a topic-based dataset consisting of the comments of all YouTube videos shared under specific domains such as STEM, LGBTQ individuals, racial minorities, or people with disabilities~\cite{chakravarthi2020hopeedi}. The detailed statistics of the HopeEDI corpus are presented in Table~\ref{tab:HopeEDIStas}\footnote{In their final version of the corpus, they did not mention the label distribution on the test set}.

Several traditional machine learning classifiers were experimented with Term Frequency-Inverse Document Frequency (TF-IDF) vectors for word uni-grams as baselines, including LR, Support Vector Machine (SVM), K-Nearest Neighbor (KNN), Decision Tree (DT), and Multinomial Naive Bayes (MNB). The DT classifier with an averaged-macro F1-score of 0.46 obtained the highest performance for the English dataset. Similarly, the results vary from 0.30 to 0.63 averaged-macro F1-scores for other languages using machine learning classifiers.

Some limitations of the existing corpora are listed as follows:
\begin{itemize}
    \item Both corpora explore only YouTube comments, whereas Twitter also is a rich source of social media texts where users can share their feelings and opinions.
    \item The low Inter-Annotator Agreement (IAA) of 0.63 for the English dataset in HopeEDI~\cite{chakravarthi2020hopeedi} corpus reveals a lack of confidence in the annotations. 
    \item Palakodety et al. (2020)~\cite{palakodety2020hope} very specifically explores only positivity and supportive comments about conflicts between India and Pakistan that reveals two issues: (i) the corpus is biased towards the conflicts between India and Pakistan about Kashmir and (ii) the concept of hope is not considered 
    \item Both the corpora are constructed for hope speech detection only as a binary classification task.
    \item The most significant criticism on both the corpora is in the annotation guidelines where they consider hope simply as a positive vibe and support, that partially may support concepts of generalized hope, optimism, and positivity. However, based on definitions of hope~\cite{snyder1991will, snyder2000handbook, snyder2002hope, rand2009hope, wong2009hope, wiles2008hope, bailey2007hope, garrard2009hope, palakodety2020hope}, they are not entirely in line with the idea of hope as having an expectation and desire of something to happen. Therefore, both corpora are more likely supportive and positive comment detection corpora.
\end{itemize}

\begin{table}[hbt!]
    \centering
    \begin{tabular}{c|cccccc}
         \bf References& \bf Languages&\bf Script&\bf Size&\bf Source&\bf Classification&\bf Avg. Macro F1 (for English)  \\
         \hline
         \cite{palakodety2020hope}&\begin{tabular}[c]{@{}c@{}}  English\\Hindi\end{tabular}&\begin{tabular}[c]{@{}c@{}} Roman\\Devanagari \end{tabular}&\begin{tabular}[c]{@{}c@{}} 2,277\\7,716 \end{tabular}& YouTube& Binary &0.78\\
         \hline
           \cite{chakravarthi2020hopeedi,  chakravarthi2022overview}&\begin{tabular}[c]{@{}c@{}}  English\\Spanish\\Tamil\\Malayalam\\Kannada\\\end{tabular}&\begin{tabular}[c]{@{}c@{}} Roman\\Code-mixed\end{tabular}&\begin{tabular}[c]{@{}c@{}} 28,424\\1,650\\17,715\\9,918\\6,176 \end{tabular}& YouTube& Binary& 0.46\\
    \end{tabular}
    \caption{Available corpora in hope speech detection}
    \label{tab:HopeSpeechDatasets}
\end{table}

\begin{table}[ht]
    \centering
    \begin{tabular}{l|ccccc}
    \multirow{2}{*}{\bf Class}& \multicolumn{5}{c}{\bf Language} \\
         
              & \bf English & \bf Spanish & \bf Tamil & \bf Malayalam & \bf Kannada\\
         \hline
         Hope & 2,234 & 660 & 7,084 &1,858  &1,909 \\
         Not Hope & 23,347&660& 8,870&6,989&3,649\\
         \hline
         Test set &2,843&330&1,761&1,071&618\\
         \hline
         Total & 28,424 & 1,650 & 17,715 & 9,918 & 6,176\\
    \end{tabular}
    \caption{Statistics of HopeEDI corpus}
    \label{tab:HopeEDIStas}
\end{table}

\subsection{Techniques for Hope Speech Detection} \label{Techniques for hope speech detection}

Chakravarthi et al. (2021, 2022)~\cite{chakravarthi2021findings, chakravarthi2022overview} held two workshops on hope speech detection on HopeEDI corpus~\cite{chakravarthi2020hopeedi} and several participants submitted their methodologies in these workshops. The statistics of corpus for all languages are given in Table~\ref{tab:HopeEDIStas}. A brief description of the works  carried out by several researchers is given below:

During the first workshop~\cite{chakravarthi-muralidaran-2021-findings}, an approach of ensembling 11 different machine learning and Neural Network models such as LR, SVM, RF, LSTM, etc., and transformers (RoBERTa, BERT, ALBERT, IndicBERT) gave averaged-weighted F1-scores of 0.93 for English comments. Similar performances were reported by a methodology consisting combination of contextualized string embedding, stacked word embeddings, and pooled document embedding with Recurrent Neural Network (RNN)~\cite{junaida2021ku_nlp}. Several other participants, explored transformers individually such as RoBERTa~\cite{mahajan2021teamuncc}, XML-R~\cite{hossain2021nlp}, XLM-RoBERTa~\cite{ziehe-etal-2021-gcdh}, XLM-RoBERTa  with TF-IDF~\cite{huang2021team}, ALBERT with  k-fold cross-validation~\cite{chen2021cs_english} and multilingual-BERT model with convolution neural networks (CNN)~\cite{dowlagar-mamidi-2021-edione}. All models scored the same averaged-weighted F1-score of 0.93.  
Among the vast transformer-based models, Balouchzahi et al. (2021)~\cite{balouchzahi-etal-2021-mucs-lt} preferred working on feature engineering for traditional machine learning. They combined char sequences with traditional word n-grams and syntactic word n-grams~\cite{sidorov2012syntactic, sidorov2014syntactic} for English texts. Their proposed approach got an averaged-weighted F1-score of 0.92 in English, a performance comparable with that of transformers.

Similarly, in the second workshop on hope speech detection~\cite{chakravarthi2022overview}, most participants preferred experimenting with different transformers and their ensembles and multilingual versions of them (e.g., BERT, Roberta, XLM, MLNet, IndicBert). They all scored around 0.87 ($\pm 0.1$) averaged-weighted F1-score on the updated version of the corpus.

Few other teams explored different learning models, such as: 
A majority voting ensemble approach of LSTM-based models, including LSTM, CNN+LSTM, and BiLSTM, gave averaged-weighted scores of 0.88 and 0.76 for English and Spanish comments~\cite{zhu2022lps} respectively.
 Gupta et al. (2022)~\cite{gupta2022iit} used TF-IDF vectors with several machine learning classifiers, namely: SVM, LR, MNB, Random Forest Classifier (RFC), and Extreme Gradient Boosting (XGB). However, their methodology was unsuccessful, with an averaged-weighted F1-score of 0.61.

The solutions explored by participants show a significant gap in studying features that affect hope speech, such as emotions~(see the definitions of hope in~\ref{Introduction}. INTRODUCTION). Moreover, few systems experimented techniques to deal with the imbalanced distribution of labels in the corpus, and the rest developed their systems overlooking these two factors.

Only Balouchzahi et al. (2022)~\cite{balouchzahi2022cic} presented a model using sequential neural networks that were trained on Linguistic Inquiry and Word Count (LIWC) psychological features~\cite{tausczik2010psychological} combined with n-gram features. They scored the third rank with averaged-weighted scores of 0.87 and 0.79 for English and Spanish texts, respectively. Also, one of the teams~\cite{gowda2022mucic} used the over-sampling technique to address the imbalanced dataset and used a 1D CNN-LSTM architecture to tackle the hope speech detection in English comments. Their methodology proved its effectiveness by securing the first rank in the competition. A summary of various techniques used for binary hope speech detection in English is presented in Table~\ref{tab:Techniuqes}.

The related studies show a considerable gap in the understanding of the concept of hope, and they often relied only on topical features and transformers. Hence, impactful features and methods need to be explored~\cite{balouchzahi2022cic}. On the other hand, the high averaged-weighted F1-scores were mainly obtained because of the larger size of the Not Hope class in the corpus (see Table~\ref{tab:HopeEDIStas}). The significant contribution of the Not Hope class in averaged-weighted F1-score caused a very high score, which resulted in an incomplete representation of hope speech detection models, as none of them gave an averaged-macro F1-score of more than 0.60~\cite{balouchzahi2022cic} in the works submitted in these shared tasks. In the most recent work~\cite{chakravarthi2022hope}, a concatenation of embedding from T5-Sentence was used to train a CNN model that outperformed all existing models with an averaged-macro F1-score of 0.75.

\begin{table}[ht]
    \centering
    \resizebox{\columnwidth}{!}{%
    \begin{tabular}{ccccc}
         \bf Reference&  \bf Model&\bf Features& \bf Avg. Weighted F1 &\bf Avg. Macro F1 \\
         \hline
         \cite{gowda2022mucic} &1D CNN-LSTM& Embedding & 0.86 & 0.55\\
         \cite{balouchzahi2022cic}& Sequential NNs& LIWC+n-grams& 0.87&0.53\\
         \cite{muti2022leaningtower}& Transformers& BERT& 0.87&0.53\\
         \cite{surana2022ginius}& Transformers&  RoBERTa  & 0.86&0.51\\
         \cite{zhu2022lps}&  LSTM, BiLSTM, BiLSTM-CNN& Embedding& 0.88&0.41\\
         \cite{bharathi2022ssncse_nlp}& Transformers& mBERT, MLNet, BERT, XLM-R, and XLM\_MLM& 0.88&0.40\\
         \cite{kumar2022soa_nlp}&LR and SVM& Char level TF-IDF&0.88&0.38\\
         \cite{chakravarthi2022hope}& CNN & T5-Sentence & 0.91 & 0.75\\
         
    \end{tabular}
    }
    \caption{Techniques for binary hope speech detection in English}
    \label{tab:Techniuqes}
\end{table}

\section{DATASET DEVELOPMENT} \label{Dataset Development}
This proposed work provides a novel dataset of tweets in English collected during the first half of 2022 to form a two-level hope speech detection task. First, the 50,000 most recent tweets were collected from Jan-2022 to Jun-2022. Then the second 50,000 tweets were collected in the same timeline using keywords related to hope (such as hope, Inshallah, aspire, believe, expect, want, wish, etc., and their different variations).
Altogether, the raw data collected from Jan-2022 to Jun-2022 consists of around 100,000 tweets, including recent tweets and tweets based on keywords.

\subsection{Data Collection and Processing} \label{Data collection and processing}

Tweepy\footnote{https://docs.tweepy.org/en/stable/api.html} API enables us to filter out the tweets based on various fields such as date, location, language, id, and other information related to the tweets and is used to collect the tweets. The date and language fields were used to filter tweets while scraping so that only English tweets belonging to the first half of 2022 were collected. Around 100,000 tweets were collected that mainly covers women's child abortion rights, black people's rights, religion, and politics domains. The collected tweets revealed that, in several cases, Tweepy could not scrape the full-text tweets, and some of the tweets were incomplete. Therefore, we removed all incomplete and duplicate tweets, which reduced the total number of tweets to less than 70,000. Further, tweets with less than 10 words were also removed, resulting in less than 50,000 tweets. Out of the remaining tweets, all retweets were removed, and only the original tweets were used, which amounted to about 23,000 tweets. These tweets were shuffled, and a sample of 10,000 tweets was randomly taken for the annotation process. However, this is not the final number of tweets, as some of the tweets were removed during the annotation process. (see~\ref{Dataset statistics}. Dataset Statistics for details.)

Inspired by Bevendorff et al. (2021)~\cite{bevendorff2021overview}, our scientific research and its outcomes are committed to the legal and ethical issues related to collecting, analyzing, and profiling social media data (based on Rangel and Rosso (2019)~\cite{rangel2019implications}). Therefore, we have blinded all usernames and URLs to \#USER\# and \#URL\#, respectively.

\subsection{Annotation guidelines} \label{Annotation guidelines}
Hope is a complex human state of mind that influences expectations but is a distinct and multifacted cognitive that is difficult to describe~\cite{eaves2016ways}. Webb (2007)~\cite{webb2007modes} believes that the experience of different types of hope highly depends on the environment and society that one may live. He argues that hope is a highly differentiated experience, with each type having different characteristics and conceptualizations. Eaves et al. (2016)~\cite{eaves2016ways} portraying hope as a mystery argues that distinguishing between the types of hope would flatten our understanding of hope in a better way.

In view of the definition of hope given by Psychologists, in the first level, we classify the tweets into one of the two classes Hope (tweets that convey a mention of hope) and Not Hope (tweets that do not convey hope). In the second level, we categorize the tweets into different types of hope based on features and characteristics such as the objective, specificity, probability of happening, nature, and emotion embedded in the tweets. The annotation guidelines used for the second level categorization of tweets are presented below, and sample tweets are given in Table~\ref{tab:samples}:

\begin{itemize}
    \item \textbf{Not Hope: }The tweet does not indicate hope, wish, desire, or future-oriented expectation.

    \item \textbf{Generalized Hope: }This kind of hope is expressed as a general hopefulness and optimism that is not directed toward any specific event or outcome~\cite{wiles2008hope, webb2007modes, ezzy2000illness, lohne2004hope, smith2005men}.

    \item \textbf{Unrealistic Hope: }The likelihood of occurrence of this kind of hope is very rare (or may not even exist). It is usually in the form of wishing for something to become true, even though the possibility of happening is remote or significantly less or even zero. Sometimes, one may hope for some irrational event/outcome out of anger, sadness, or depression. This type of hope is meaningless and is without any reasoning\cite{webb2007modes, eliott2002discursive,links1994breaking}.

    \item \textbf{Realistic Hope: }This kind of hope is about expecting something reasonable, meaningful, and possible thing to happen (there is every possibility that this will happen). It includes any hope for a reasonable or probable outcome in terms of regular and expected outcomes~\cite{wiles2008hope, eaves2016ways, garrard2009hope}.

\end{itemize}

\begin{table}[ht]
    \centering
    \resizebox{\columnwidth}{!}{%
    \begin{tabular}{l|l}
        \bf Categories & \bf Samples  \\ \hline
         \multirow{4}{*}{Not Hope}& 
                                \begin{tabular}[c]{@{}l@{}}\#USER\# \#USER\# It’s also common sense. You don’t need to pray or be religious for that.\end{tabular}\\ 
                                
                                \cline{2-2}
                                 &\begin{tabular}[c]{@{}l@{}}This is still a single entry point. It doesn't allow for older students to enter the program, even if they have the skills or desire to.\\ It is still a program where the credentials are given based on the verified classroom instruction time rather than skills\end{tabular}\\ 
                                 
                                 \cline{2-2}
                                 &\begin{tabular}[c]{@{}l@{}} Yes that's what I do but every cycle is different and it's not reliable, so I cannot count on that to work\end{tabular}\\ 
                                 
                                 \cline{2-2}
                                 &\begin{tabular}[c]{@{}l@{}} \#USER\# You just can’t get out of your own way, can you? \end{tabular}\\
        \hline
         \multirow{4}{*}{Generalized Hope} & 
                                \begin{tabular}[c]{@{}l@{}}This is so tragic, I hope everyone injured makes a speedy recovery and offer my condolences for those who lost a loved one\\today in this horrible accident. ???? \#URL\#\end{tabular}\\ 
                                
                                \cline{2-2}
                                 &\begin{tabular}[c]{@{}l@{}} Happiest birthday to our dearest Min Hyuk ???? We hope that you'll enjoy this special day of yours. We wish you all the best\\because you deserve it. We are so proud of you! \end{tabular}\\ 
                                 
                                 \cline{2-2}
                                 &\begin{tabular}[c]{@{}l@{}} Congratulations \#USER\# bro for completing, my allah god bless u ahead a good and bright future inshallah... \#URL\# \end{tabular}\\ 
                                 
                                 \cline{2-2}
                                 &\begin{tabular}[c]{@{}l@{}} \#USER\# You don’t know how much i adore and talk to my frinds about you (in a good way). I love you, your style and your\\energy and i pray for your recovery. Stay strong love \end{tabular}\\ 
                                            
        \hline
         \multirow{4}{*}{Unrealistic Hope} & 
                                \begin{tabular}[c]{@{}l@{}} \#USER\# “I wish I had a tail as big as a kite, just like Libra~!" \end{tabular}\\ 
                                
                                \cline{2-2}
                                 &\begin{tabular}[c]{@{}l@{}} i see cheesecake factory and think of oomf's desire of jungkook being a cheesecake factory waitor \end{tabular}\\ 
                                 
                                 \cline{2-2}
                                 &\begin{tabular}[c]{@{}l@{}} \#USER\# damn man i'm here just tryna be a nice friend and dis is how u treat me i hope u choke with cold spaghetti \end{tabular}\\ 
                                 
                                 \cline{2-2}
                                 &\begin{tabular}[c]{@{}l@{}} \#USER\# \#USER\# \#USER\# U n ur 4amily will never see peace bcos of your religion mindset. Inshallah Amen \end{tabular}\\

        \hline
         \multirow{4}{*}{Realistic Hope} & 
                                \begin{tabular}[c]{@{}l@{}} \#USER\# I think this is a good project with clear future prospects.  I hope this project continues as we expect because it is\\supported by a good team. \end{tabular}\\ 
                                
                                \cline{2-2}
                                 &\begin{tabular}[c]{@{}l@{}} I have spent a lot of time with my cousins and I’m forever grateful but I am making way back to traveling and I’m hoping to\\bring my family with me to places eventually but we need to get back in our modes. We all are lagging haha \end{tabular}\\ 
                                 
                                 \cline{2-2}
                                 &\begin{tabular}[c]{@{}l@{}} It was really hard exam but fortunately I studied a little last night and I expect to pass the test \end{tabular}\\ 
                                 
                                 \cline{2-2}
                                 &\begin{tabular}[c]{@{}l@{}} You have been working all your life \#USER\# definitely you can buy a car soon \end{tabular}\\ 
    \end{tabular}
    }
    \caption{Sample tweets from the dataset}
    \label{tab:samples}
\end{table}

\subsection{Annotator selection} \label{Annotator selection}
As identifying the different types of hope was one of the main challenges during the annotation of the dataset, a strict method employed to select the qualified annotators for the annotation task, which is described below: 

\begin{enumerate}
    \item 200 sample tweets, along with detailed annotation guidelines, were given to eight candidate annotators.
    \item After receiving the labeled samples from annotators, each annotator’s labels were analyzed, and the five best annotators who gave better performance were selected for the next round.
    \item One-to-one meetings and interviews were conducted with each one of these five annotators to clear their confusion on annotating the sample tweets to make sure they completely understood the task (without directly pointing to the annotation of specific tweets, we instead discussed the labels where they had the most confusion along with a complete discussion on guidelines).
    \item The five selected annotators were given the next 200 samples for annotations, and their performances were supervised.
    \item Finally, among the five annotators, only three who eventually improved their annotations and had a better inter-annotator agreement on the sample tweets were selected to annotate the tweets to create the hope speech dataset.

\end{enumerate}

\subsection{Annotation Procedure} \label{Annotation procedure}
The detailed annotation guidelines (see \ref{Annotation guidelines}. Annotation Guidelines) and samples were provided to the three selected annotators (see \ref{Annotator selection}. Annotator Selection) to assist in the annotation process for creating a hope speech detection dataset. All three annotators (A1, A2, and A3) had a good knowledge of English and at least a post-graduate degree. Two annotators were from a computer science background and were familiar with NLP tasks and machine learning concepts, and the other was from a psychology background.

To monitor the annotation process, the dataset was divided into three batches. For each batch, a meeting was conducted with annotators to analyze the annotation process and discuss the challenges they were facing. Further, the annotators were asked to follow the  procedure given below along with guidelines:

\begin{enumerate}
    \item \textbf{First level annotation: }If there is no sign of hope in the tweet, mark the tweet as Not Hope; if the tweet conveys a sign of hope, mark the tweet as Hope for the first level and check the type of hope for the second level.

    \item \textbf{Second level annotation: }If a tweet was annotated as Hope in the first level and if the tweet is about some general expectation for a better future (based on the definition of Generalized hope in “\ref{Annotation guidelines}. Annotation guidelines") that tweet should be labeled as Generalized Hope.  Further, suppose the tweet is directed toward a specific event/outcome. In that case, the annotators should check whether it is a Realistic or Unrealistic hope (based on the definition of realistic and unrealistic hopes in “\ref{Annotation guidelines}. Annotation guidelines") and label the tweet accordingly.
    
\end{enumerate}

\begin{figure}[ht]
\centering
    \includegraphics[width=1\textwidth]{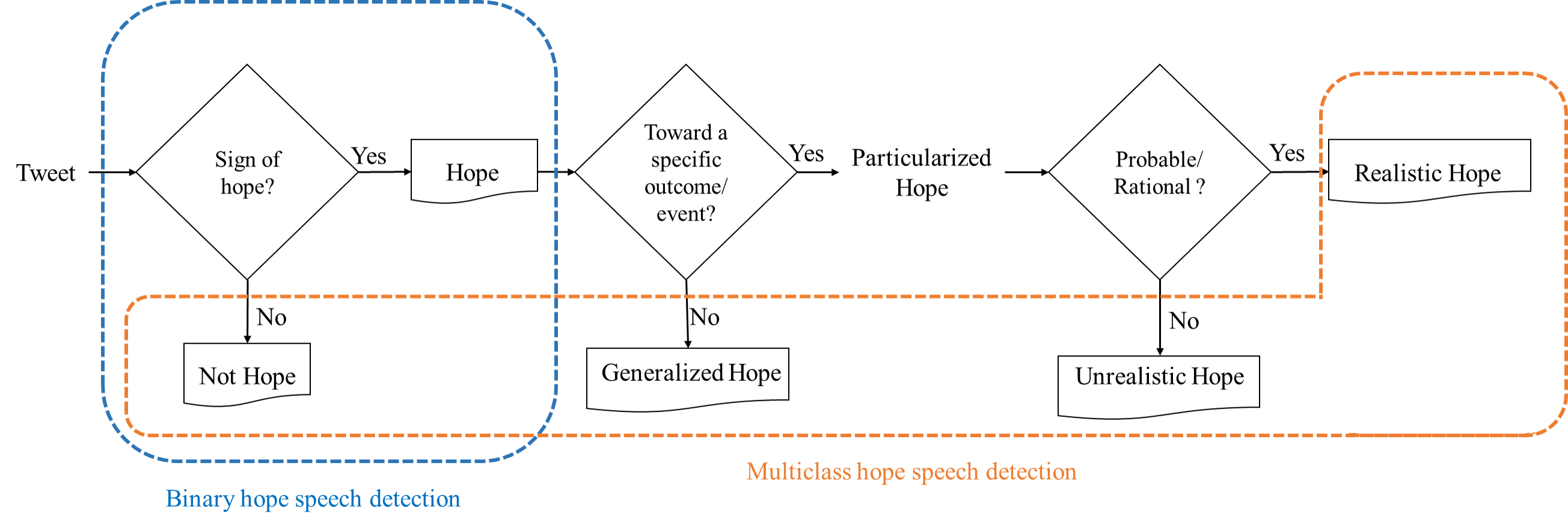}
\caption{Workflow of annotation procedure}
\label{fig:ann}
\end{figure}

A simple graphical representation of the annotation procedure is illustrated in Figure~\ref{fig:ann}.

\subsection{Inter-Annotator Agreement} \label{Inter-annotator agreement}

Inter-Annotator Agreement (IAA) measures the agreement between annotators while considering the possibility of chance agreement. A Cohen’s Kappa Coefficient~\cite{cohen1960coefficient} of 85\% for binary hope speech dataset and a Fleiss' Kappa Score~\cite{falotico2015fleiss} of 82\% for multiclass hope speech dataset indicate the strength of the datasets, which is the result of a strict annotation process.

\subsection{ Statistics of the Dataset} \label{Dataset statistics}
\begin{figure}[ht]
\centering
    \includegraphics[width=1\textwidth]{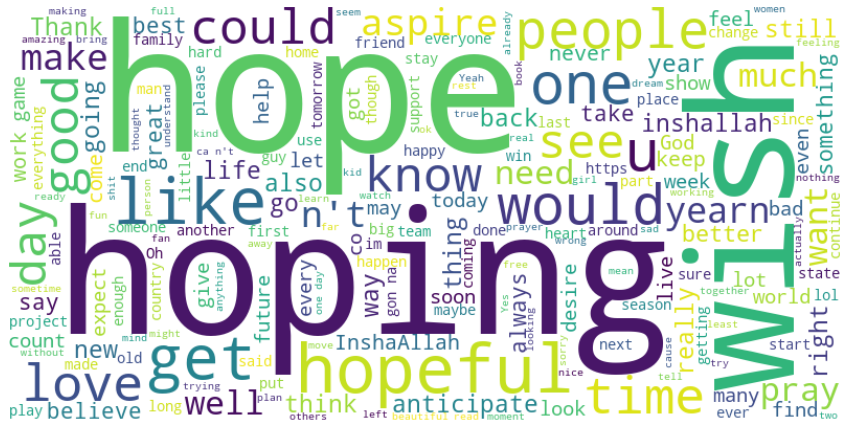}
\caption{Word cloud of hope keywords in the dataset}
\label{fig:HopeWC}
\end{figure}

\begin{figure}[ht]
\centering
    \includegraphics[width=0.6\textwidth]{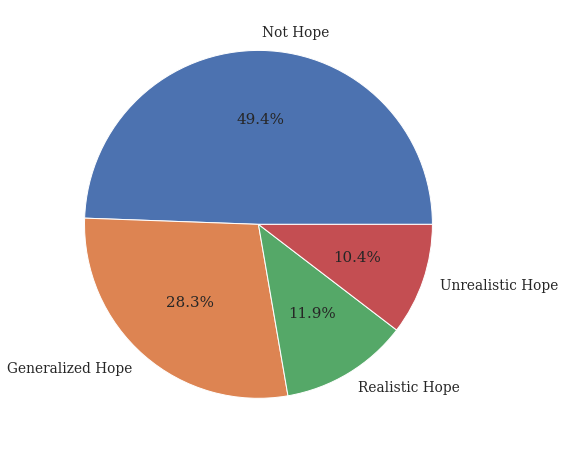}
\caption{Label distribution in the multiclass hope speech detection subtask}
\label{fig:DataDist}
\end{figure}

Despite filtering the tweets by language in the initial stage, some tweets included texts in multi-script such as Persian, Urdu, and Hindi with English. The annotators removed such tweets amounting to 8,256 tweets, out of which 4,175 tweets are labeled as Hope and the remaining 4,081 tweets labeled as  Not Hope; for the first level (binary) classification which shows an entirely balanced distribution for the first level classification. The characteristics of the dataset and statistics for the second level (multiclass) classification are presented in Table~\ref{tab:CorpStas}. The dataset has an average of 32.97 words, and the vocabulary size is 40,410. The observations show that unrealistic hopes usually were expressed in fewer words (average number of words = 28.84).
In contrast, tweets with more words (average number of words = 34.68) usually belong to the Not Hope category. A word cloud of keywords in Hope tweets is presented in Figure~\ref{fig:HopeWC}. Label distribution of the dataset shown in Figure~\ref{fig:DataDist} illustrates that roughly half of the tweets were labeled as different types of Hope, and the other half were identified as Not Hope tweets.

\begin{table}[ht]
\centering

\begin{tabular}{lllclcl}
\bf Class & \bf Tweets  & \bf Words & \bf Avg. Words  & \bf Characters  & \bf Avg. Characters & \bf Vocab\\
\hline
Not Hope & 4081 & 141538 & 34.68 & 795016 & 194.80 & 26294\\
Generalized Hope  & 2335 &  73677 & 31.55 & 405383 & 173.61 & 15003\\
Realistic Hope & 982 & 32316 & 32.90 & 179372 & 182.65 & 8359\\
Unrealistic Hope & 858 & 24746 & 28.84 & 130893 & 152.55 & 6502\\
\hline
\bf Total & 8256 & 272277 & 32.97 & 1510664 & 182.97 & 40410\\
\end{tabular}

\caption{\label{tab:CorpStas}Statistics of the dataset}
\end{table}

\section{BENCHMARKS  } \label{BENCHMARKS}
A set of traditional machine learning, deep learning, and transformer models were experimented on the proposed dataset to set the benchmarks and to analyze the dataset and reliability of the annotations.

\subsection{Preprocessing} \label{Preprocessing}
A uniform preprocessing method was used for all the experiments. Contractions are converted  to standard forms (e.g., “you're" and “should've" are converted to “you are" and “should have") using a dictionary and stop words are  removed (except for transformers experiments) followed by removing Unicodes (e.g., “x89û", “x89ûò"),  HTML/XML character entity references (e.g., “\&gt;", “\&lt;", “\&amp;"), user mentions, URLs, and non-alphabet and non-numeric characters (including emojis and non-Roman alphabets).  Further, mis-spaced words were corrected using WordNet, and regular expressions (e.g., “ChicagoArea" and “socialnews" were converted to “Chicago Area" and “social news"), and the text was lowercased.

\subsection{Traditional Machine Learning Models} \label{Traditional machine learning models}
Eight traditional machine learning classifiers, namely: LR, SVM with Radial Basis Function (RBF), and linear kernels, RFC, XGB, AdaBoost, and Catboost, are used for the hope speech detection task. All the classifiers are used with default parameters and are trained on the TF-IDF vectors of word uni-grams.

\subsection{Deep Learning Models} \label{Deep learning models}
Two deep learning models, namely: Convolutional Neural Network (CNN) and Bidirectional Long Short-Term Memory (BiLSTM), are trained separately with Global Vectors for Word Representation (GloVe) and FastText embeddings.
A Keras\footnote{https://keras.io/} tokenizer was fitted on the texts from the dataset and was used to convert all texts to sequences. The maximum length of the sequence is set to the maximum length of tweets in the dataset, and all sequences were padded to the same length.  Vectors are obtained from the word embedding matrix for each tweet, and then the input sequences are created and fed to the deep learning models. The parameters used for both deep learning models are presented in Table~\ref{tab:ParamDL}, and the models are trained on 20 epochs for each fold.

\begin{table}[ht]
\centering

\begin{tabular}{lcc}
 \bf Parameters & \bf CNN & \bf BiLSTM \\
\hline
    Epochs & 20 per fold & 20 per fold\\
    Optimizer & Adam & Adam\\
    loss & categorical crossentropy & categorical crossentropy\\
    filter sizes & [1,2,3,5] & -\\
    number of filters & 36 & -\\
    embedding size& 300 &300\\
    lr & 0.001 & 0.001\\
    Dropout & 0.1 &0.1\\
    Activation & softmax &softmax\\
    
\end{tabular}
                       
\caption{\label{tab:ParamDL}Parameters for deep learning models}
\end{table}

\subsection{Transformers} \label{Transformers}

Many NLP researches are inspired by the methods which have already proven their effectiveness in other fields to apply these concepts to NLP, such as Computer Vision (CV), including transfer learning approaches wherein the knowledge of pre-trained models on ImageNet is transferred to target tasks~\cite{fazlourrahman2022coffitt}.

HuggingFace platform enables researchers to explore various open source tools, codes, and technologies in various domains, including, Image Processing, NLP, and other data science works. Several transformers from HuggingFace\footnote{https://huggingface.co/} platform were employed and fine-tuned using SimpleTransformers\footnote{https://simpletransformers.ai/} library that provides a user-friendly API to initialize, train (or fine-tune) and evaluate a task-specific transformer model from HuggingFace. Six transformer-based models, namely: BERT~\cite{kenton2019bert}, ALBERT~\cite{lan2019albert}, Roberta~\cite{liu2019roberta}, DistilBERT~\cite{sanh2019distilbert}, XLNet~\cite{yang2019xlnet}, and ELECTRA~\cite{clark2019electra}, are used in this paper. Maximum sequence length and learning rate were set to 100 and 3e-5, respectively, and the rest of the parameters were set as default. All models were fine-tuned for 15 epochs in 5 folds.

\section{RESULTS} \label{Results}
Precision, Recall, and F1-scores, along with micro, macro, and weighted are the most widely used metrics and the averaging methods used for evaluating classification models' performance. These averaging methods are differentiated based on how they consider the contribution of classes or samples in the final averaged score. The micro-averaged method favors a balanced dataset, which considers the proportion of correctly classified samples out of all samples (all samples have equal contribution to the final averaged score). In the case of an unbalanced dataset, classes with more samples will have a larger impact on the final score. The macro-averaged method treats all classes equally, regardless of the number of samples, which is an advantage over the micro-averaged method for imbalanced datasets. On the other hand, the weighted-averaged method calculates the scores by taking the mean of all per-class scores, considering the size of each class (number of samples).   
As the proposed multiclass hope speech detection dataset has an imbalanced distribution of labels,  macro-averaged and weighted-averaged scores are more suitable than micro-averaged scores. Five-fold cross-validation was used for all learning models, and the mean scores of all folds were used as final scores for the evaluation.

\subsection{Traditional machine learning models}
Table~\ref{tab:ResML} presents the overall results obtained by the machine learning models using TF-IDF of word n-grams. The results show that the machine learning models have exhibited competitive and similar performance. However, LR and CatBoost classifiers performed better in terms of averaged-macro F1-score and averaged-weighted F1-score for binary and multiclass classification, respectively, compared to the rest of the machine learning models. General observation of the performance of the machine learning classifiers revealed that boosting classifiers (AdaBoost, CatBoost, and XGB) performed better than the rest of the classifiers for the multiclass hope speech detection task.

\begin{table}[ht]
\centering

\begin{tabular}{l|ccc|ccc|c}
 \multirow{2}{*}{\bf Model} & \multicolumn{3}{c}{\bf Avg. Weighted Scores} & \multicolumn{3}{c}{\bf Avg. Macro Scores} & \multirow{2}{*}{\bf Accuracy} \\
        &Precision&Recall&F1-score&Precision&Recall&F1-score& \\

\hline
\multicolumn{8}{c}{\bf Binary Hope Speech Detection}\\
\hline
LR &0.80&0.80&0.80&0.80&0.80&0.80&0.80 \\
SVM (rbf) &0.80&0.79&0.79&0.80&0.79&0.79&0.79 \\
SVM (linear) &0.79&0.79&0.79&0.79&0.79&0.79&0.79 \\
RFC &0.79&0.79&0.79&0.79&0.79&0.79&0.79  \\
AdaBoost &0.79&0.79&0.79&0.79&0.79&0.79&0.79  \\
XGB &0.80&0.79&0.79&0.80&0.79&0.79&0.79  \\
CatBoost &0.80&0.79&0.79&0.80&0.79&0.79&0.79 \\
\hline
\multicolumn{8}{c}{\bf Multiclass Hope Speech Detection}\\
\hline
LR &0.64&0.66&0.62&0.61&0.48&0.50&0.66 \\
SVM (rbf) &0.63&0.65&0.60&0.61&0.45&0.45&0.65 \\
SVM (linear) &0.64&0.66&0.64&0.60&0.50&0.52&0.66 \\
RFC &0.64&0.67&0.64&0.60&0.51&0.53&0.67  \\
AdaBoost &0.62&0.64&0.63&0.54&0.52&0.53&0.64  \\
XGB &0.63&0.66&0.64&0.58&0.51&0.53&0.66  \\
CatBoost &0.64&0.66&0.64&0.59&0.52&0.54&0.66  \\
\end{tabular}

\caption{\label{tab:ResML}Results for machine learning models}
\end{table}

\subsection{Deep learning models}
The performance of deep learning models using GloVe and FasText word embeddings are shown in Table~\ref{tab:ResDL}. The observations show that deep learning models with FastText embeddings clearly outperformed the models using GloVe embeddings for the binary classification task. For the multiclass task, the BiLSTM model using FastText embeddings has improved results compared to CNN with FastText and also the machine learning models. However, the best results using deep learning models were obtained using GloVe embeddings for the multiclass task. CNN and BiLSTM models with GloVe embeddings achieved almost identical averaged scores with only a slight difference in averaged-macro precision.

\begin{table}[ht]
\centering
\resizebox{\columnwidth}{!}{%
\begin{tabular}{l|c|ccc|ccc|c}
 \multirow{2}{*}{\bf Word Embedding}&\multirow{2}{*}{\bf Model} & \multicolumn{3}{c}{\bf Avg. Weighted Scores} & \multicolumn{3}{c}{\bf Avg. Macro Scores} & \multirow{2}{*}{\bf Accuracy} \\
                                 &&Precision&Recall&F1-score&Precision&Recall&F1-score& \\

\hline
\multicolumn{9}{c}{\bf Binary Hope Speech Detection}\\
\hline
\bf \multirow{2}{*}{FastText}& CNN &0.78&0.78&0.78&0.78&0.78&0.78&0.78 \\
\bf                              & BiLSTM &0.79&0.79&0.79&0.79&0.79&0.79&0.79 \\
\cline{2-9}
\bf \multirow{2}{*}{GloVe}& CNN &0.82&0.82&0.82&0.82&0.82&0.82&0.82 \\
\bf                              &BiLSTM &0.82&0.82&0.82&0.82&0.82&0.82&0.82 \\

\hline
\multicolumn{9}{c}{\bf Multiclass Hope Speech Detection}\\
\hline
\bf \multirow{2}{*}{FastText}& CNN &0.63&0.64&0.64&0.56&0.53&0.54&0.64 \\
\bf                              &BiLSTM &0.66&0.67&0.66&0.60&0.56&0.57&0.67  \\
\cline{2-9}
\bf \multirow{2}{*}{GloVe}   & CNN &0.70&0.70&0.70&0.64&0.60&0.61&0.70 \\
\bf                              & BiLSTM &0.70&0.70&0.70&0.63&0.60&0.61&0.70 \\

\end{tabular}
}
\caption{\label{tab:ResDL}Results for deep learning models}
\end{table}

\subsection{Transformer models}

Table~\ref{tab:ResTransformers} provides the performance obtained by different transformer models. Most transformer models obtained an averaged-weighted F1-score of higher than 0.83 for binary and higher than 0.70 for multiclass hope speech detection, which illustrates the strength of transformers over the rest of the learning models.  For the binary classification task, models using BERT, Roberta, and XLNet obtained the highest results with averaged-macro F1-scores of 0.85. On the other hand, for the multiclass task BERT~\cite{kenton2019bert} model outperformed the rest of the transformers with averaged-weighted and averaged-macro F1-scores of 0.77 and 0.72, respectively. 
A comparison of the performances of the different learning approaches shows remarkable differences between the transformers and other learning models.

\begin{table}[ht]
\centering
\resizebox{\columnwidth}{!}{%
\begin{tabular}{l|ccc|ccc|c}
 \multirow{2}{*}{\bf Transformers} & \multicolumn{3}{c}{\bf Avg. Weighted Scores} & \multicolumn{3}{c}{\bf Avg. Macro Scores} & \multirow{2}{*}{\bf Accuracy} \\
&Precision&Recall&F1-score&Precision&Recall&F1-score& \\
\hline
\multicolumn{8}{c}{\bf Binary Hope Speech Detection}\\
\hline
bert-base-uncased & 0.85& 0.85& 0.85& 0.85& 0.85& 0.85& 0.85 \\
roberta-base &0.85& 0.85& 0.85& 0.85& 0.85& 0.85& 0.85 \\
albert-base-v2 &0.83&0.83&0.83&0.83&0.83&0.83&0.83 \\
xlnet-base-cased &0.86&0.85&0.85&0.86&0.85&0.85&0.85  \\
google/electra-base &0.84&0.84&0.84&0.84&0.84&0.84&0.84  \\
distilbert-base-uncased &0.85&0.85&0.85&0.86&0.85&0.85&0.85  \\

\hline
\multicolumn{8}{c}{\bf Multiclass Hope Speech Detection}\\
\hline
\bf bert-base-uncased & \bf 0.78& \bf 0.77& \bf 0.77& \bf 0.71& \bf 0.72& \bf 0.72& \bf0.77 \\
roberta-base &0.77&0.76&0.76&0.69&0.71&0.70&0.76 \\
albert-base-v2 &0.75&0.75&0.75&0.69&0.67&0.68&0.75 \\
xlnet-base-cased &0.76&0.75&0.76&0.69&0.71&0.69&0.74  \\
google/electra-base &0.75&0.74&0.74&0.67&0.69&0.68&0.74  \\
distilbert-base-uncased &0.76&0.76&0.76&0.70&0.70&0.70&0.76  \\

\end{tabular}
}
\caption{\label{tab:ResTransformers}Results for transformers}
\end{table}

\section{ERROR ANALYSIS} \label{Error Analysis}
\begin{figure*}[ht]
    \centering
    \begin{subfigure}[t]{0.5\textwidth}
        \centering
        \includegraphics[height=1.5in]{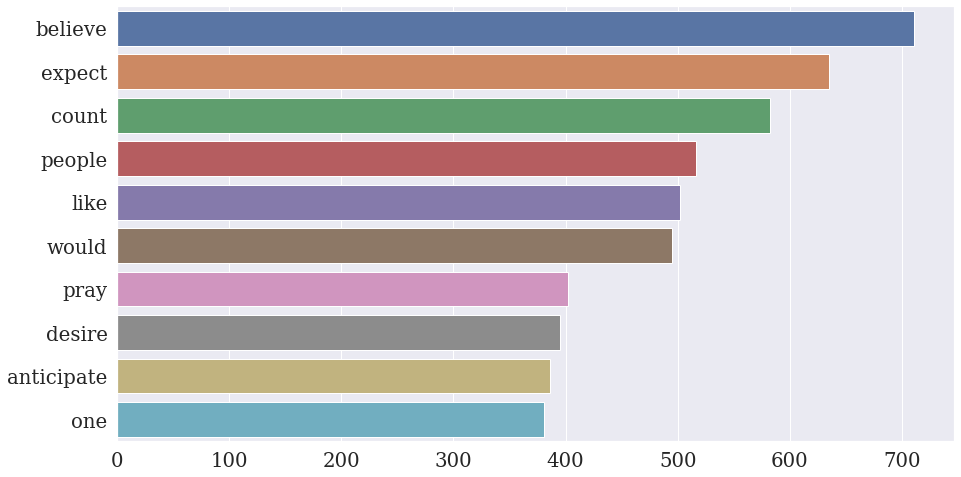}
        \caption{Not Hope}
    \end{subfigure}%
    ~ 
    \begin{subfigure}[t]{0.5\textwidth}
        \centering
        \includegraphics[height=1.5in]{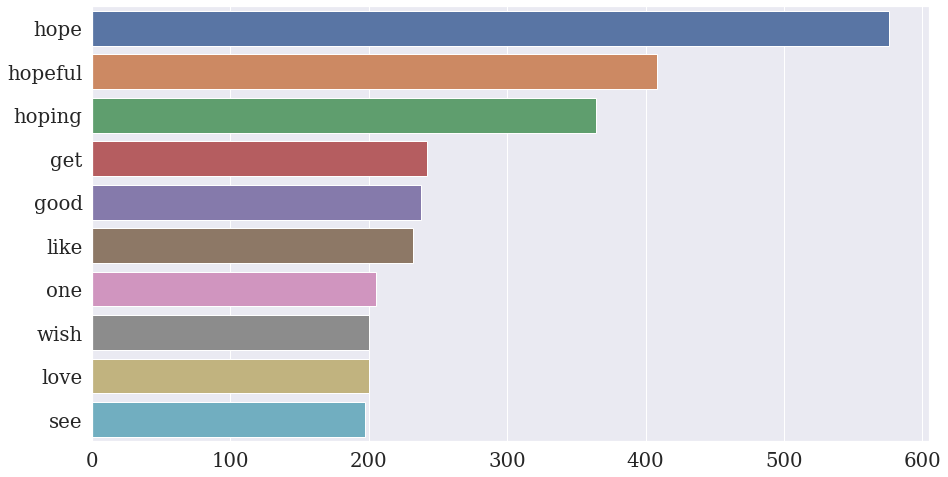}
        \caption{Generalized Hope}
    \end{subfigure}
    ~
    \begin{subfigure}[t]{0.5\textwidth}
        \centering
        \includegraphics[height=1.5in]{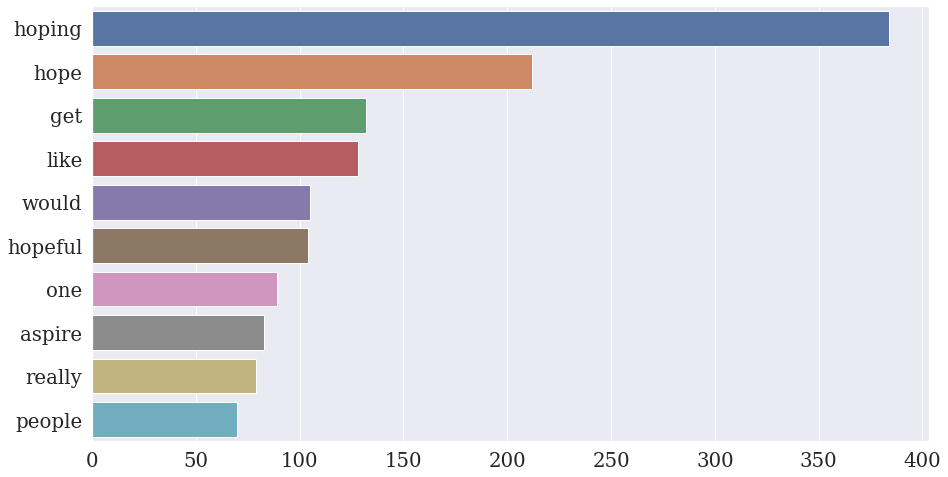}
        \caption{Realistic Hope}
    \end{subfigure}%
    ~
     \begin{subfigure}[t]{0.5\textwidth}
        \centering
        \includegraphics[height=1.5in]{Figures/Common_unigrams_GHope.png}
        \caption{Unrealistic Hope}
    \end{subfigure}
    \caption{Most common uni-grams (after preprocessing and removing stopwords) for each class (multiclass)}
\label{fig:unigrams}
\end{figure*}

Table~\ref{tab:bestPerforming} shows the results of the best-performing models in each learning approach. 
The observations reveal that simple machine learning classifiers were behind other learning models in multiclass hope speech detection. Figure~\ref{fig:unigrams} justifies the lower results for machine learning models using uni-grams. As can be seen, there are many overlaps among uni-gram features of different classes. These overlaps are more significant for the three Hope categories. This shows that simple uni-grams, despite their excellent results on binary classification task, are not suitable features for multiclass hope speech detection task. 

On the other hand, deep learning and transformers models employ more complicated features and structures with attention (for transformers~\cite{vaswani2017attention}) resulted in higher performance~\footnote{Since the focus in this paper was building a benchmark dataset, we skipped experimenting with more complicated features and architectures in the current paper}. It is observed that models utilizing more information from text, such as context, significantly have higher results (e.g., transformers and deep learning models with GloVe have higher performance than the deep learning models using FastText embedding for the multiclass task).

\begin{table}[ht]
    \centering
    \begin{tabular}{l|lcc}

         Model&Learning approach& Averaged-weighted F1 & Averaged-macro F1\\
         \hline
          \multicolumn{4}{c}{\bf Binary Hope Speech Detection}\\
        \hline
        LR& Machine learning& 0.80 & 0.80\\
         BiLSTM and CNN both with FastText& Deep learning & 0.82 & 0.82\\
         \bf BERT, Roberta, and XLNet& \bf Transformers& \bf 0.85 & \bf 0.85\\
         
         \hline
        \multicolumn{4}{c}{\bf Multiclass Hope Speech Detection}\\
        \hline
         CatBoost& Machine learning& 0.64 & 0.54\\
         CNN+GloVe& Deep learning & 0.70 & 0.61\\
         \bf BERT& \bf Transformers& \bf 0.77 & \bf 0.72\\
    \end{tabular}
    \caption{Best performing models in each learning approach}
    \label{tab:bestPerforming}
\end{table}

\begin{figure*}[t!]
    \centering
    \begin{subfigure}[t]{0.4\textwidth}
        \centering
        \includegraphics[height=2.3in]{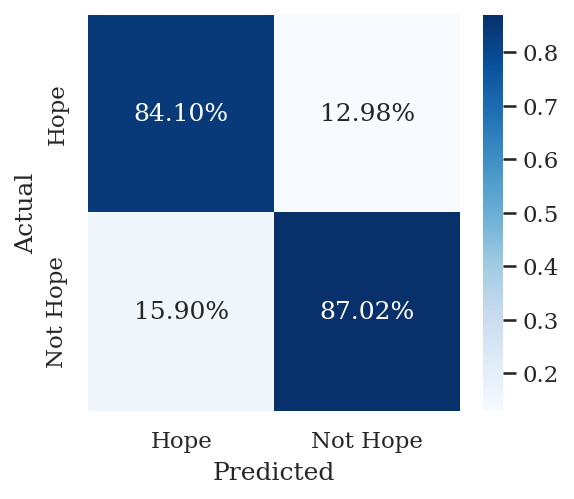}
        \caption{Binary hope speech classification}
    \end{subfigure}%
     ~
    \begin{subfigure}[t]{0.4\textwidth}
        \centering
        \includegraphics[height=2.3in]{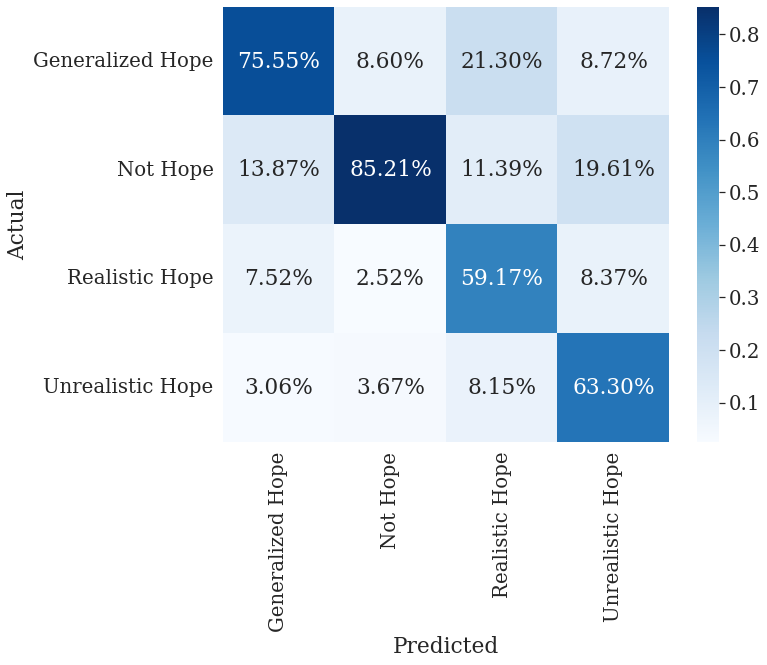}
        \caption{Multiclass hope speech classification}
    \end{subfigure}
    
    \caption{Confusion matrix for the BERT model}
\label{fig:BertCM}
\end{figure*}

\begin{figure*}[ht]
    \centering
    \begin{subfigure}[t]{0.6\textwidth}
        \includegraphics[height=2.1in]{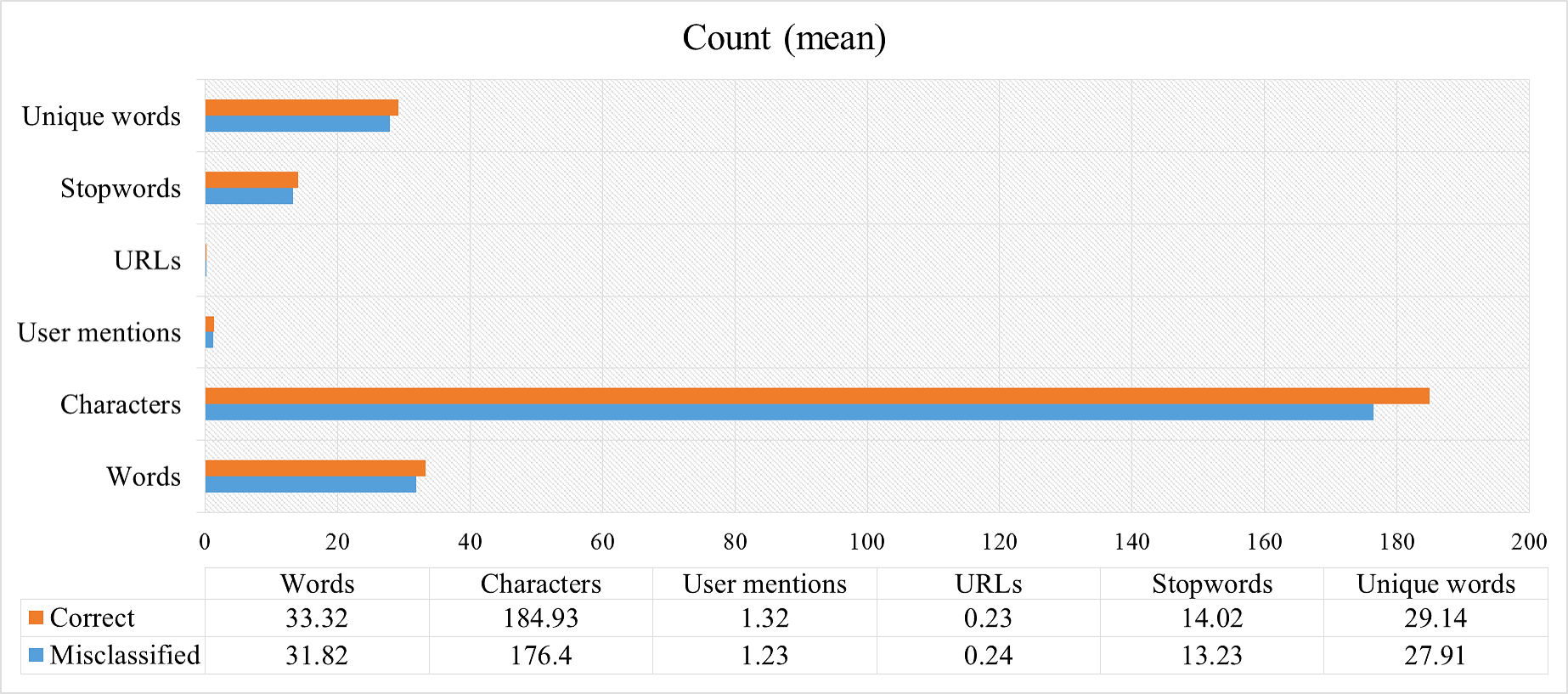}
        \caption{Without preprocessing}
    \end{subfigure}%
     
    \begin{subfigure}[t]{0.6\textwidth}
        \includegraphics[height=2.1in]{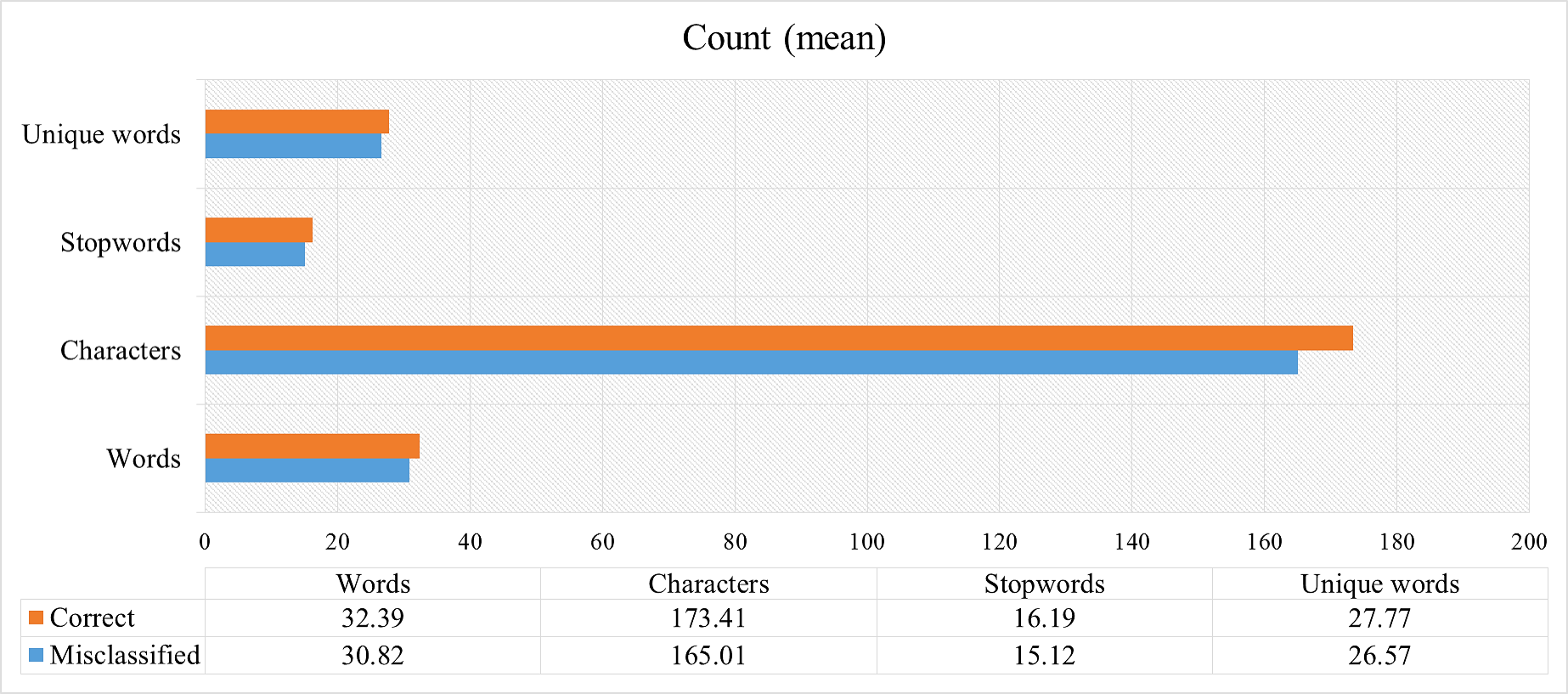}
        \caption{With preprocessing}
    \end{subfigure}
    
    \caption{Statistics for the misclassified tweets (multiclass)}
\label{fig:MissClassStatis}
\end{figure*}

\begin{figure*}[t!]
    \centering
    \begin{subfigure}[t]{0.5\textwidth}
        \centering
        \includegraphics[height=1.5in]{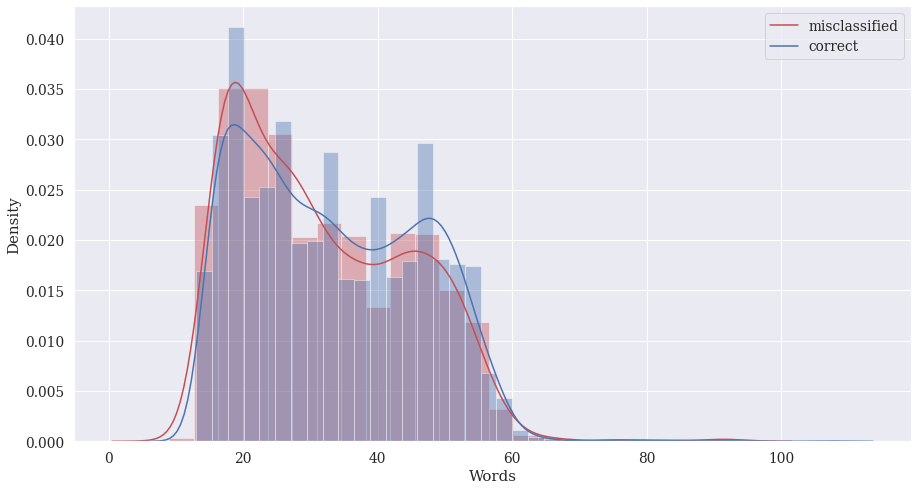}
        \caption{Word count Without preprocessing}
    \end{subfigure}%
    ~ 
    \begin{subfigure}[t]{0.5\textwidth}
        \centering
        \includegraphics[height=1.5in]{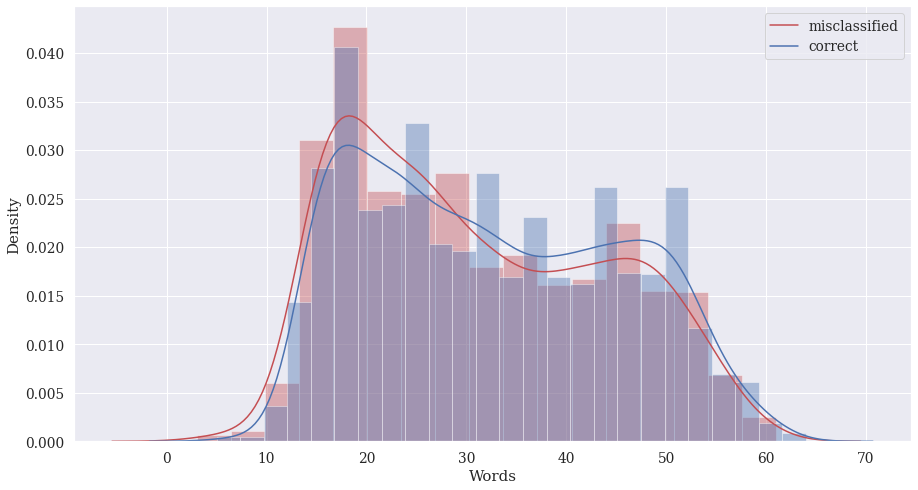}
        \caption{Word count With preprocessing}
    \end{subfigure}
    ~
    \begin{subfigure}[t]{0.5\textwidth}
        \centering
        \includegraphics[height=1.5in]{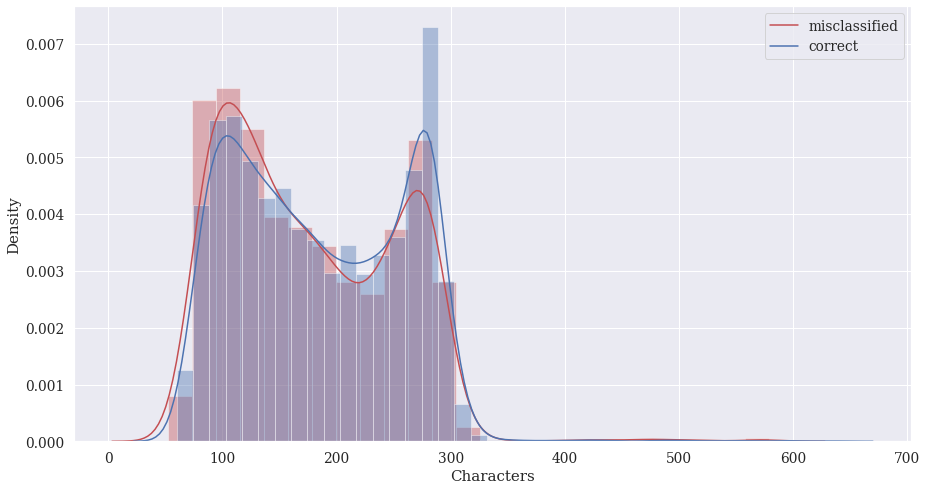}
        \caption{Character count Without preprocessing}
    \end{subfigure}%
    ~ 
    \begin{subfigure}[t]{0.5\textwidth}
        \centering
        \includegraphics[height=1.5in]{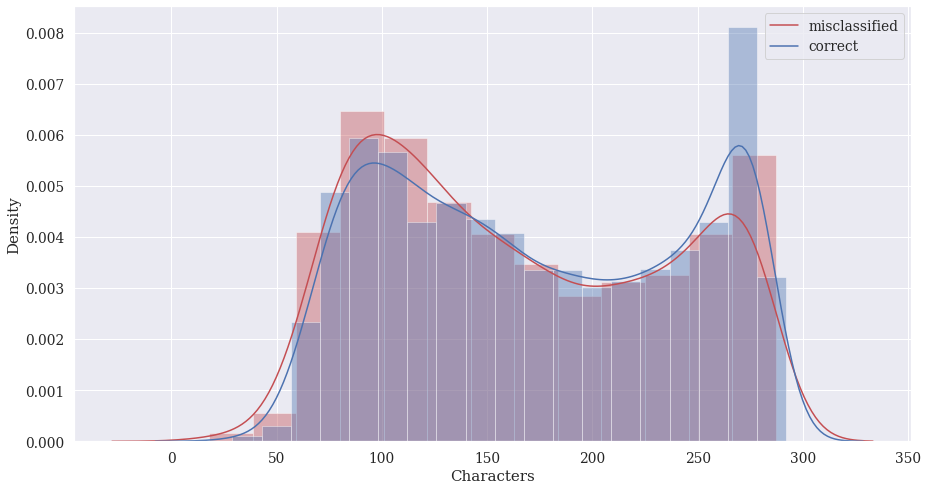}
        \caption{Character count With preprocessing}
    \end{subfigure}
        ~
    \begin{subfigure}[t]{0.5\textwidth}
        \centering
        \includegraphics[height=1.5in]{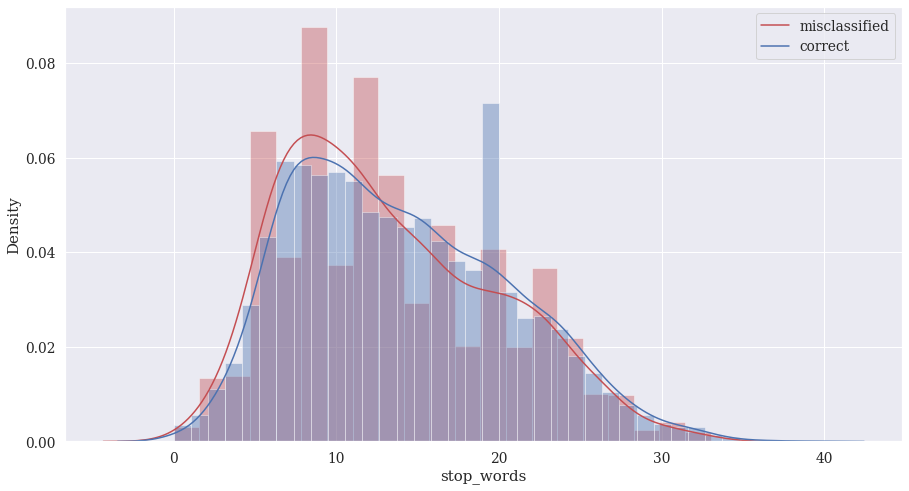}
        \caption{Stopword count Without preprocessing}
    \end{subfigure}%
    ~ 
    \begin{subfigure}[t]{0.5\textwidth}
        \centering
        \includegraphics[height=1.5in]{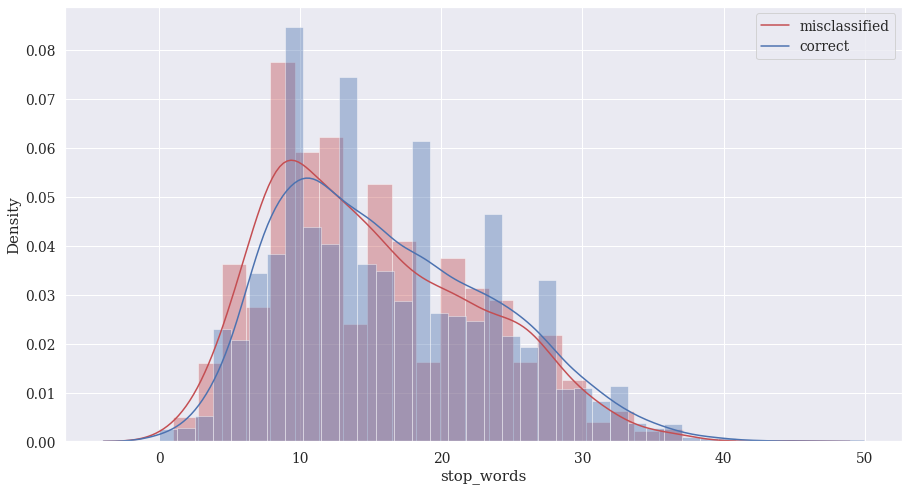}
        \caption{Stopword count With preprocessing}
    \end{subfigure}
    
        ~
    \begin{subfigure}[t]{0.5\textwidth}
        \centering
        \includegraphics[height=1.5in]{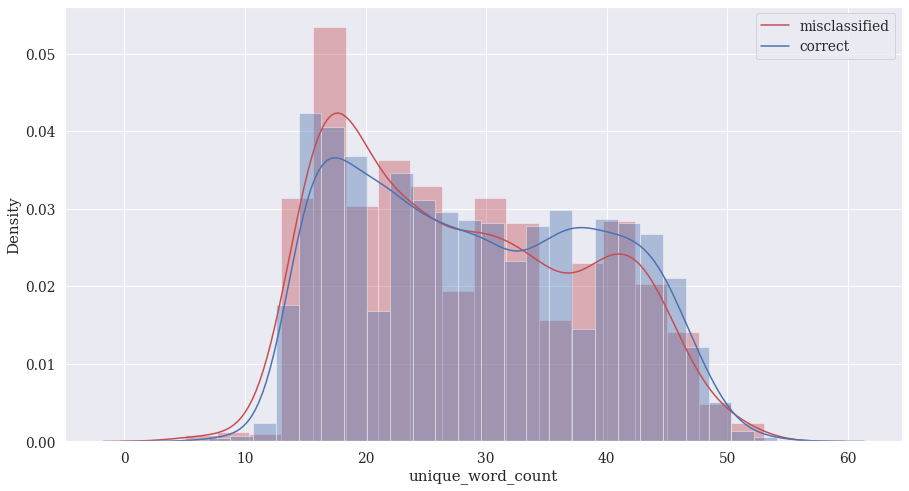}
        \caption{Unique word count Without preprocessing}
    \end{subfigure}%
    ~ 
    \begin{subfigure}[t]{0.5\textwidth}
        \centering
        \includegraphics[height=1.5in]{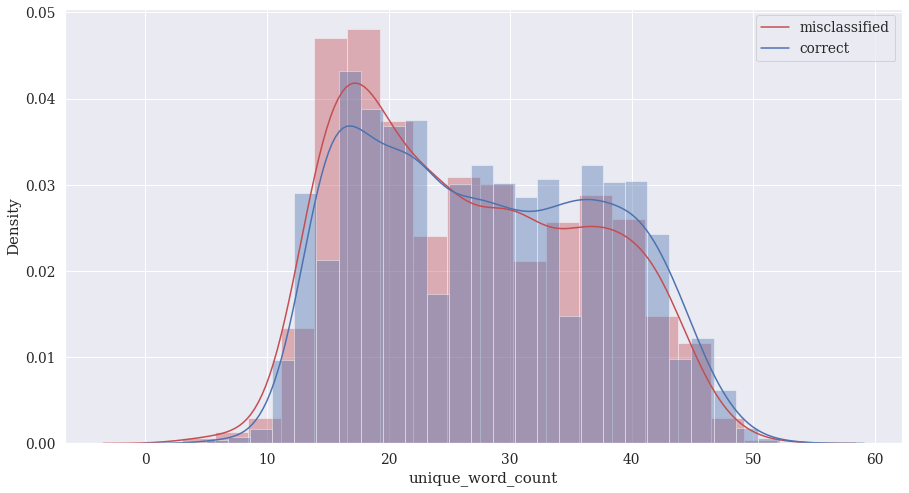}
        \caption{Unique word count With preprocessing}
    \end{subfigure}
    
    \caption{Detailed statistics for the misclassified tweets (multiclass)}
\label{fig:MissClassStatisDetailed}
\end{figure*}

Table~\ref{tab:clsreport} presents classwise scores for both binary and multiclass classification for BERT model. The confusion matrix of the BERT model on the proposed dataset shown in Figure~\ref{fig:BertCM} illustrates that in most cases, the model successfully identified classes in the binary task and Not Hope tweets in the multiclass task. Further, the model performed well on the Generalized hope but obtained lower performances for the realistic and unrealistic hope categories compared to the other labels, which was in line with our expectations. Very generally, having fewer samples of these two categories produced less accuracy in predictions. The misclassified samples for the BERT model for the multiclass task were analyzed, and observations are illustrated in Figure~\ref{fig:MissClassStatis}. The statistics of misclassified and correctly identified samples show that misclassified samples were usually shorter and had less information than correctly identified samples.
Regarding preprocessed samples used for the training and predictions, correctly identified samples had an average number of 32.39 words and 173.41  characters per tweet. In contrast, misclassified tweets were shorter, with an average of 30.82 words and 165.01  characters per tweet. A similar average count for user mentions and URLs for all tweets shows that they did not significantly impact misclassifying them. Detailed statistics of the misclassified and correctly identified samples are presented in Figure~\ref{fig:MissClassStatisDetailed}.

\begin{table}[ht]
\centering

\begin{tabular}{l|ccc|c}
\bf Categories & Precision&Recall&F1-score&support \\
\hline
\multicolumn{5}{c}{\bf Binary Hope Speech Detection}\\
\hline
Not Hope &0.84 &0.88 &0.86&816  \\
Hope & 0.87&0.83&0.85&835 \\
\hline
\multicolumn{5}{c}{\bf Multiclass Hope Speech Detection}\\
\hline
Not Hope &0.85 &0.85 &0.85&817  \\
Generalized Hope & 0.76&0.72&0.74&467 \\
Unrealistic Hope &0.63&0.64&0.64&196 \\
Realistic Hope &0.59&0.65&0.62&172 \\
\end{tabular}

\caption{\label{tab:clsreport}Classwise scores for the BERT model}
\end{table}

\section{DISCUSSION} \label{DISCUSSION}
\begin{figure}[ht]
\centering
\begin{subfigure}[t]{0.5\textwidth}
        \centering
        \includegraphics[height=1.7in]{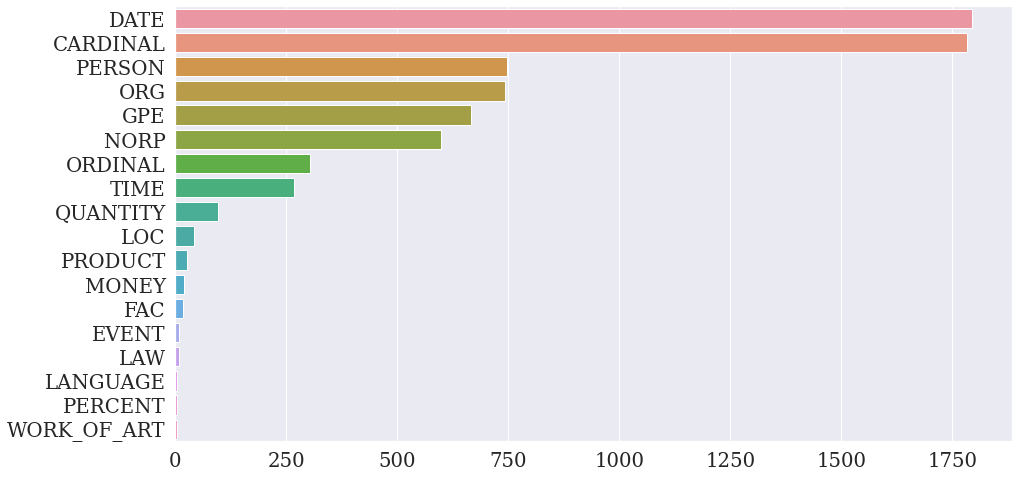}
        \caption{NER tags in the Dataset}
    \end{subfigure}%
    ~ 
    \begin{subfigure}[t]{0.5\textwidth}
        \centering
        \includegraphics[height=1.7in]{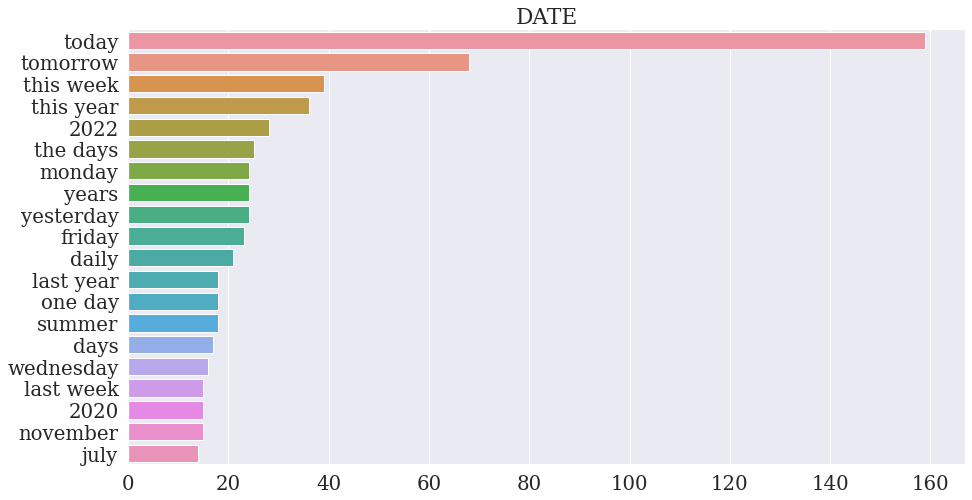}
        \caption{Samples for DATE tag}
    \end{subfigure}
\caption{NER tags in the dataset}
\label{fig:NEsInDS}
\end{figure}

\begin{figure}[ht]
\centering
    \includegraphics[width=0.6\textwidth]{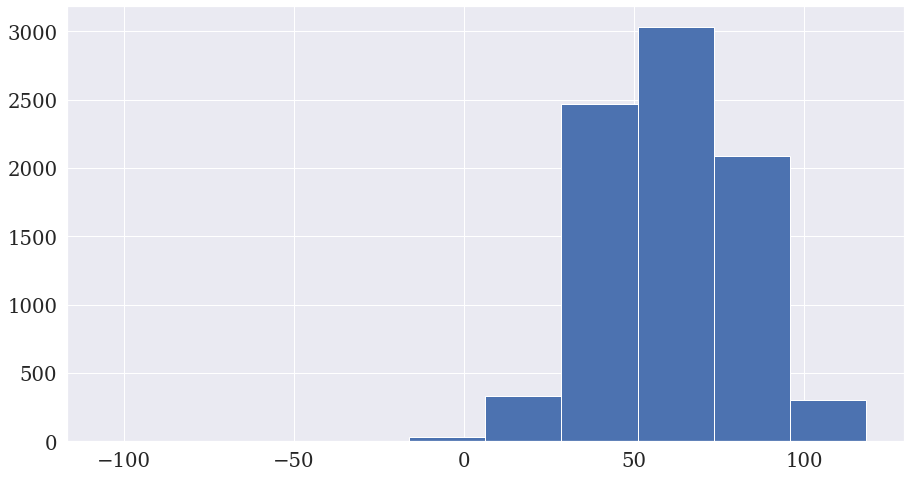}
\caption{Readability and complexity of the dataset}
\label{fig:readability}
\end{figure}

The proposed dataset has several characteristics and limitations, as summarized below:

\begin{enumerate}
    \item The dataset includes only the original tweets as retweets were discarded in the early stage of processing the tweets. This is based on the assumption that retweets may not have sufficient information to contribute to the classification task.

    \item Despite having a strict and robust annotation process, the annotators faced several challenges during annotations, especially for annotating unrealistic hope. Sometimes, the tweet may represent hope but wishing for something harmful (e.g., I hope all of them will be killed \#USER\# \#USER\# \#USER\#); in such cases, some annotators believe that these tweets support the definition of having a desire to something happen, and also they were future-oriented; even though in the form of anger and sadness. On the other hand, some annotators believed that hope is always associated with expecting a favorable future (where the first group thought the hoper might think of death as a positive outcome for himself). Therefore, we gave freedom to label the tweets based on the annotators' understanding of tweets. However, in most cases, these types of tweets were labeled as unrealistic hope since they are irrational, with no or less likelihood of occurrence.  In several cases, annotators believed some tweets represented regret rather than a wish or hope and labeled them as not hope since they are not future-oriented  (e.g., \#USER\# my paternal grandparent died long ago, I wish I had spent more time with them).

    \item The proposed dataset tackles the concept of hope based on the definition given by Psychologists (see \ref{Introduction} INTRODUCTION) rather than only taking positivity and optimism as hope. Considering Figure~\ref{fig:unigrams} and~\ref{fig:NEsInDS} the definition of hope as a future-oriented expectation, desire, wish and want is justified. The Spacy NER tagger was used for named-entity extraction in the dataset, and Figure~\ref{fig:NEsInDS} represents the NER tags over the dataset. It is observed that DATE (this tag indicates date and time) is the most frequent tag, and the samples show that they are often future-oriented.

    \item The readability of tweets was evaluated using textstat\footnote{https://pypi.org/project/textstat/} library. It is an easy tool to determine readability, complexity, and grade level. The result of the readability analysis shown in Figure~\ref{fig:readability} illustrates that most tweets scored a readability score of more than 50, which means that the dataset is relatively readable.
    
    \item The low IAA score of HopeEDI~\cite{chakravarthi2020hopeedi} (IAA score of 63\% for binary hope speech detection in English YouTube comments) proved the difficulty of the annotation process for hope speech detection. Therefore, our strict annotator selection and strong guidelines resulted in an IAA of 0.85\% for binary and 82\% in multiclass classifications in the proposed dataset.
    \item The performance of the models for both subtasks is entirely in line with IAA scores which reveals the validity of our dataset. Further, it shows that annotators and learning models struggled to classify 20\% ($\pm$ 5\%) of tweets in the dataset.
\end{enumerate}

\section{CONCLUSION AND FUTURE WORK}\label{CONCLUSION}
This paper describes the two-level dataset for the task of hope speech detection in English tweets. To the best of our knowledge, this is the first-ever dataset that tackles the hope speech detection task with the actual concept of hope and as a multiclass classification task. In particular, we performed a strict annotator selection process to select the most suitable annotators for this dataset. We provided them with robust and detailed annotation guidelines that resulted in an IAA score of 0.85\% for binary and 82\% in multiclass classification tasks. Initially, we collected more than 100,000 recent tweets and tweets based on keywords (such as hope, Inshallah, aspire, believe, expect, want, wish, etc., and their different variations) tweeted during the first half of the year 2022. The tweets were commonly in the domains of women's child abortion rights, black people's rights, religion, and politics. Eventually, the final annotated dataset had 8,256 tweets, out of which 4,175 tweets were labeled as Hope, and the other 4,081 were labeled as Not hope classes in the first level. The hope tweets were further categorized into one of three classes: Generalized Hope, Realistic Hope, and Unrealistic hope. 

We benchmarked the dataset with a set of baselines using various learning approaches. The baselines include machine learning models trained using TF-IDF word uni-grams, CNN and BiLSTM with GloVe and FastText embeddings, and several transformers. We found that the multiclass hope speech task is challenging. The machine learning models with simple n-grams performed well on binary hope speech detection but struggled to identify different types of hope. On the other side, neural network models with embedding and language models such as BiLSTM, CNN, and especially transformers gave a better performance for multiclass hope speech detection task. However, the performance of traditional machine learning models employing more complicated features and feature engineering methods to empower them is still questionable. Among all our baselines, the bert-base-uncased transformer with an averaged-macro F1-score of 0.85 and 0.72 for binary and multiclass hope speech detection obtained the highest results.

We hope our dataset will enable NLP researchers to further explore hope as a new psychological task on the Internet and social network platforms. In addition, similar to other social media analysis tasks, such as emotion analysis and depression detection, hope speech detection can be used in a wide range of NLP applications, such as public health, mental health, anxiety, and human reaction detection for human behavior analysis.
In the future, we plan to increase the dataset size to observe how increasing the samples of realistic and unrealistic hope would affect the performance of learning models. Further, we would like to extend our work in various dimensions, such as hope-related topics, different languages, and different social media platforms. We would also like to explore emotion features (e.g.,  LIWC features ) and techniques to overcome imbalance data distribution issues on the performance of machine learning models.

\section*{ACKNOWLEDGMENTS}

The work was done with partial support from the Mexican Government through the grant A1-S-47854 of CONACYT, Mexico, grants 20220852 and 20220859 of the Secretaría de Investigación y Posgrado of the Instituto Politécnico Nacional, Mexico. The authors thank the CONACYT for the computing resources brought to them through the Plataforma de Aprendizaje Profundo para Tecnologías del Lenguaje of the Laboratorio de Supercómputo of the INAOE, Mexico and acknowledge the support of Microsoft through the Microsoft Latin America PhD Award.

\section*{FUNDING}
We acknowledge support from the NAACL Regional Americas Fund in the construction of our dataset.

\bibliographystyle{unsrt}  
\bibliography{references}

\end{document}